\pdfoutput=1

\documentclass[11pt]{article}

\usepackage[]{acl}

\usepackage{amsmath}
\usepackage{amssymb}
\usepackage[utf8]{inputenc}
\usepackage{setspace}
\definecolor{lightgreen}{rgb}{0.56, 0.93, 0.56}

\usepackage{hhline}

\usepackage{times}
\usepackage{latexsym}

\usepackage[T1]{fontenc}

\usepackage[utf8]{inputenc}

\usepackage{microtype}

\usepackage{tabularx}
\usepackage{multirow}
\usepackage{booktabs}
\usepackage{graphicx}
\usepackage{listings}
\usepackage{color}
\usepackage{xcolor}
\usepackage[T1]{fontenc}
\usepackage[utf8]{inputenc}
\usepackage{listings}
\usepackage{amssymb}
\usepackage[belowskip=-7pt,aboveskip=8pt]{caption}

\usepackage{color}

\definecolor{darkred}{rgb}{0.6,0.0,0.0}
\definecolor{darkgreen}{rgb}{0,0.50,0}
\definecolor{lightblue}{rgb}{0.0,0.42,0.91}
\definecolor{orange}{rgb}{0.99,0.48,0.13}
\definecolor{grass}{rgb}{0.18,0.80,0.18}
\definecolor{pink}{rgb}{0.97,0.15,0.45}
\definecolor{codegreen}{rgb}{0,0.6,0}
\definecolor{codegray}{rgb}{0.5,0.5,0.5}
\definecolor{codepurple}{rgb}{0.58,0,0.82}
\definecolor{backcolour}{rgb}{0.95,0.95,0.92}
\definecolor{keywordcolor}{rgb}{0.7, 0.1, 0.1}   
\definecolor{tacticcolor}{rgb}{0.0, 0.1, 0.6}    
\definecolor{commentcolor}{rgb}{0.4, 0.4, 0.4}   
\definecolor{symbolcolor}{rgb}{0.0, 0.1, 0.6}    
\definecolor{sortcolor}{rgb}{0.1, 0.5, 0.1}      
\definecolor{attributecolor}{rgb}{0.7, 0.1, 0.1} 

\lstdefinestyle{mystyle}{
  frame=single,
  basicstyle=\setstretch{1.5}\ttfamily\footnotesize
  backgroundcolor=\color{backcolour}, 
  commentstyle=\color{codegreen},
  commentstyle=\color{darkgreen}\slshape,
  keywordstyle=\color{blue},
  stringstyle=\color{darkred},
  numberstyle=\tiny\color{codegray},
  emphstyle=\color{pink}\underbar,
  morekeywords={Verify, Question},
  escapeinside={(*@}{@*)},
  breakatwhitespace=false,         
  breaklines=true,                 
  captionpos=b,                    
  keepspaces=true,                    
  numbersep=5pt,                  
  showspaces=false,                
  showstringspaces=false,
  showtabs=false,                  
  tabsize=2
}
\DeclareUnicodeCharacter{2200}{\forall}
\DeclareUnicodeCharacter{2227}{\logicand}
\DeclareUnicodeCharacter{2192}{\rightarrow}

\title{LeanReasoner: Boosting Complex Logical Reasoning with Lean}

\author{Dongwei Jiang$^\heartsuit$ \quad Marcio Fonseca$^\diamondsuit$ \quad Shay B. Cohen$^\diamondsuit$  \quad   \\
  $^\heartsuit$ Johns Hopkins University $^\diamondsuit$ University of Edinburgh\\
  \texttt{djiang21@jhu.edu\quad m.fonseca@ed.ac.uk\quad scohen@inf.ed.ac.uk}
}

\newenvironment{itemizesquish}[2]{\begin{list}{\labelitemi}{\setlength{\itemsep}{#1}\setlength{\labelwidth}{#2}\setlength{\leftmargin}{\labelwidth}\addtolength{\leftmargin}{\labelsep}}}{\end{list}}

\begin{document}
\maketitle
\begin{abstract}
Large language models (LLMs) often struggle with complex logical reasoning due to logical inconsistencies and the inherent difficulty of such reasoning. 
We use Lean, a theorem proving framework, to address these challenges. By formalizing logical reasoning problems into theorems within Lean, we can solve them by proving or disproving the corresponding theorems. 
This method reduces the risk of logical inconsistencies with the help of Lean's symbolic solver. It also enhances our ability to treat complex reasoning tasks by using Lean's extensive library of theorem proofs. 
Our method achieves state-of-the-art performance on the FOLIO dataset and achieves performance near this level on ProofWriter. Notably, these results were accomplished by fine-tuning on fewer than 100 in-domain samples for each dataset.\footnote{Our code and data is available at \url{https://github.com/Some-random/theorem-proving-reasoning}.}
\end{abstract}

\section{Introduction}

Logical reasoning, a bedrock of intelligence and a core capability of humans, has been a challenging issue for machine learning systems for a long time.
LLMs, despite their impressive abilities to understand and generate natural language, often fall short when dealing with complex logical reasoning tasks. They frequently suffer from logical inconsistencies, where the model hallucinates and makes statements
not grounded in premises, leading to spurious results \cite{greedy_reasoner, content_effects}.

Recent advances in AI have adopted a structured approach to tackle these reasoning problems by splitting them into symbolic formalization and problem-solving \cite{yueya, logic-lm, satlm}. Specifically, the formalization step is often handled by a large language model, while problem-solving is handled by an off-the-shelf symbolic solver. In this approach, symbolic solvers essentially act as a rigorous checkpoint, ensuring that the model outputs align with logical rules, thereby mitigating the issue of logic inconsistency. In these approaches, solvers may range from being completely deterministic, like SymPy \cite{yueya}, or relying on a combination of heuristics and basic machine learning techniques, as is the case with Pyke \cite{logic-lm} and Z3 \cite{satlm, Z3}. While this approach successfully addresses hallucinations, it still struggles with more complex problems. 

As a powerful theorem prover and a versatile programming language, Lean \cite{Lean} presents a compelling solution to connect symbolic solvers with linguistic resources. Much like symbolic solvers, Lean has a strict checking system that ensures each reasoning step is certified.
What distinguishes it, however, is its additional functionality as a programming language developed specifically for theorem proving. Every day, a substantial amount of code is written in Lean, capturing reasoning ``nuggets'' with step-by-step rationals that are useful for training LLMs. A few recent studies have already tapped into Lean for mathematical theorem-proving tasks \cite{curriculum, PACT, hypertree}, showing its potential in tackling difficult reasoning challenges.

In this paper, we propose LeanReasoner, a Lean-based framework to tackle logical reasoning problems.
We use LLMs to formalize natural language context into Lean and fine-tune a custom model on these problems using a modest amount of data annotated ourselves.  
As we use LLMs to dynamically generate solutions within the Lean environment, our approach stands in stark contrast to the static, pre-defined solution-finding methods of LogicLM \cite{logic-lm}, which only rely on traditional techniques like forward and backward chaining, and SATLM \cite{satlm}, which operates within the Z3 environment using a suite of predetermined algorithms and heuristics. The adaptive nature of LLMs as a solution-finding tool allows our system to evolve continuously, harnessing a vast array of reasoning data and information.



\begin{figure*}[t!]
\centering
\includegraphics[ width=0.985\textwidth]{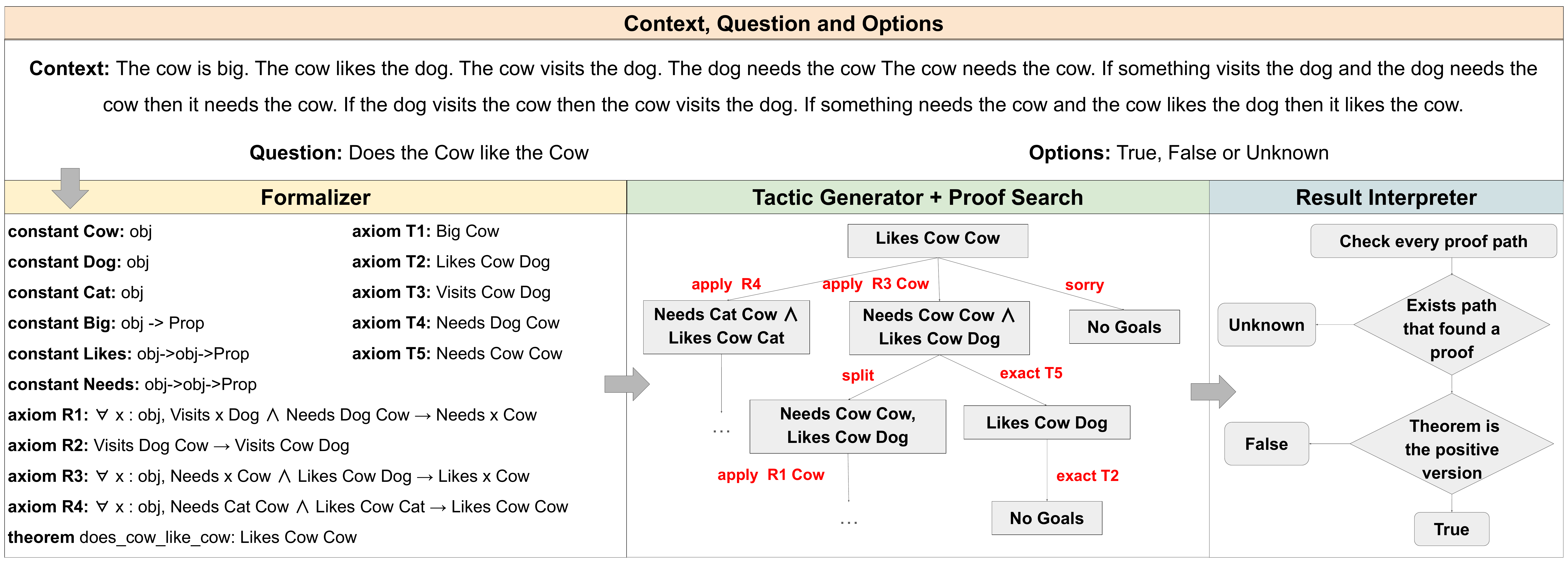} 
\caption{\textbf{An overview of our approach.} The natural language context is first processed by the ``formalizer''. It then advances to the proof search stage, where all the tactics (in red) generated by the ``tactic generator'' are used to manipulate goals. Finally, the outcome is interpreted by the ``result interpreter''.}
\label{structure}
\end{figure*}

Our contributions in this paper are three-fold.
\begin{itemizesquish}{-0.3em}{0.5em}
\item To our knowledge, this is the first attempt to use Lean, traditionally associated with mathematical theorem proving, for natural language logical reasoning. This effort highlights a possible intersection between mathematical theorem proving and logical reasoning.
\item Our research revealed that incorporating pretraining data from mathematical theorem proofs enhances the development of a more effective solver for logical reasoning compared to previous techniques. Additionally, this approach enabled us to achieve SOTA results on FOLIO.
\item We make available the training data accumulated in this research, comprising 100 Lean-formalized logic reasoning problems from ProofWriter, along with 27 analogous formalizations from FOLIO. The corresponding proofs in Lean are also included.
\end{itemizesquish}

\section{Problem Definition and Notation}

The task we aim to solve is logical reasoning, taking the form of multi-choice questions given a natural language context. The answer to the question can be logically deduced based on the context. 
The framework we use for solving the problem is Lean.\footnote{\url{https://leanprover.github.io/}.}
Lean is an open-source theorem-proving programming language with vibrant
community support. Its current base includes over 100,000 theorems and 1,000,000 lines of code.\footnote{\url{https://en.wikipedia.org/wiki/Lean_(proof_assistant)}.}
We use Lean as a generic theorem prover, outside of mathematics.

The task and our solution to it, consist of the following components:

\begin{itemizesquish}{-0.3em}{0.5em}

\item \textbf{Context}, which is composed of natural language utterances, composing a set of rules and facts. For example:
\emph{Hudson is a cat}, \emph{all cats are animals}, and \emph{cats often meow}.

\item \textbf{Question}, which denotes the posed question. For example, \emph{Does Hudson often meow?}

\item \textbf{Options} is a set of available answers (discrete categories) from which an answer can be chosen.
For example, \emph{True}, \emph{False} or \emph{Unknown}.

\item \textbf{Formalized context} refers to the representation of context in Lean. For example, the formalized context for our example would be: \emph{axiom A1 is\_cat Hudson}, \emph{axiom A2 $\forall x$, is\_cat $x$ $\rightarrow$ is\_animal $x$} and \emph{axiom A3 $\forall x$, is\_cat $x$ $\rightarrow$ often\_meow $x$}.

\item \textbf{Formalized question}: Given that Lean operates as a theorem prover, questions are transformed into dual theorems: one asserting the positive stance and the other negating it. For the given example, the formalized questions would be: \emph{Theorem hudson\_often\_meows: often\_meow\; Hudson} and \emph{Theorem not\_hudson\_often\_meows: $\neg$ often\_meow\; Hudson}.

\item \textbf{Goal}: In the context of proving theorems with Lean, a "goal" is a logical statement that needs to be proven true, given a set of axioms and rules. When we set out to answer a question using the Lean prover, this question becomes our root goal. At that point, we can apply various instructions in Lean to simplify or break down this primary goal and generate intermediate goals.

For instance, using our earlier examples, if the root goal is proving \emph{Theorem hudson\_often\_meows: often\_meow\; Hudson}, an intermediate goal might be proving that \emph{Hudson is a cat}. We aim to resolve each intermediate goal using our provided context, gradually working our way towards proving the root goal. Once all intermediate goals are resolved, we have effectively proven our root goal, and the proof search concludes successfully.

\item \textbf{Tactics} are the instructions in the Lean theorem proving language that are used to manipulate goals to obtain a proof for a given goal. For example, \emph{apply A3 Hudson} is a tactic that uses modus ponens on the \textbf{Goal} \emph{often\_meow Hudson} and transforms it to a new \textbf{Goal} \emph{is\_cat Hudson}

\end{itemizesquish}

A diagram of these components and the relations between them is depicted in Figure \ref{structure}. 
This procedure is framed within the language of the Lean theorem prover as a goal-satisfying process.

\section{LeanReasoner}
\label{section:method}

Our framework, LeanReasoner, is composed of four main components: a \emph{formalizer}, a \emph{tactic generator}, a \emph{proof search} mechanism, and a \emph{result interpreter}. The formalizer converts the context and question to formalized context and formalized question. The tactic generator then generates tactics based on premises extracted from the formalized context. The proof search mechanism oversees tactic execution and goal expansion. The result interpreter analyses the output of the proof search and identifies the correct answer in the options. In this section, we detail each of those components.

\subsection{Formalizer}
As formalizers, we used OpenAI models text-davinci-003 (GPT-3) and GPT-4 \cite{gpt4}. For text-davinci-003, we followed the same prompting approach as Logic-LM \cite{logic-lm} to separate task specifications and problems, thereby enabling the model to continue with the task of formalization through next-token-prediction. For GPT-4, we used similar prompts but included the task specification in the system prompt. 

There is no automatic way to assert all the entities, relationships, and constraints of the context have been captured by the formalized result. However, the syntax of the formalized result can be checked by Lean. Because correct syntax is a prerequisite for downstream theorem proving, if an error is encountered during compilation, we provide the error message generated by Lean along with the faulty formalization and ask the formalizer to regenerate the result. We further manually inspect the formalizer in \S\ref{section:results}. We note that we take a strict approach, and if the formalizer fails more than once, then the problem is counted as not being correctly solved.

\subsection{Tactic Generator}
The model we used for tactic generation is ReProver \cite{leandojo}. This model contains two parts: a retriever that employs retrieval mechanisms to explicitly select premises when provided with the current goal, and a generator that generates tactics using the goal and the retrieved premises.

The division of the problem-solving task into premise selection and tactic generation simplifies the process and facilitates easier troubleshooting. It isolates the source of potential issues, be it in the premise selection or the tactic generation, thus reducing the complexity of the problem. Also, this division of responsibilities eases the burden on the tactic generator. Choosing the right premise with numerous distractions is challenging, especially in logical reasoning problems when several options might seem promising for the current step but will not ultimately lead to the desired goal.

The premise retrieval component of our process draws from the Dense Passage Retriever (DPR) \cite{densepassage}. Provided with a 
goal $g$ as the query and a set of candidate premises $P$, it generates a ranked list of $m$ premises from $P$. In DPR, both $g$ and $P$ are treated as raw texts that are embedded in a vector space. We then retrieve the top $m$ premises that maximize the cosine similarity between the goal and the premise. For tactic generation, we use a standard sequence-to-sequence model. The goal and the premises are concatenated together as a string to generate new tactics.

As a baseline, we also prompt GPT-4 to generate proofs. For cases when the chosen theorem to prove aligns with the answer (say the chosen theorem is the positive stance of the question and the answer is \textsc{Yes}), we present GPT-4 with the correct proof as part of the prompt. Conversely, if the answer does not align with the chosen theorem or the answer is \textsc{Unknown}, the formalized theorem is unprovable. In those cases, we still encourage the model to engage in step-by-step reasoning, even though it will eventually hit a roadblock. An example of the prompt to GPT-4 can be found in Appendix \ref{sec:proofwriterprompt}.

\subsection{Proof Search}

The proof search module controls the overall search process that selects tactics and maintains states during proof construction. Essentially, the goal of the search method is to build a proof tree that incrementally evolves the goal through tactic invocations. This approach was first introduced in GPT-F \cite{gpt-f}. LeanDoJo \cite{leandojo}, a recently released framework that enables interaction with Lean programmatically, subsequently provided an implementation of this method, which we use for our study.

As a reference, the middle part of \autoref{structure} provides a practical illustration of this process. Starting from the root goal, for each given proof goal, we explore 64 possible tactics. All goals are maintained in a priority queue and expanded based on cumulative log probabilities of the goal. The cumulative log probability is defined as the summation of the log probabilities of the tactics that brought us to the goal from the root. This implies that we tend to expand those goals where our generative model has the highest global confidence. 

To enhance search efficiency and circumvent potential loops, we have incorporated a mechanism that stops the expansion of a node $N$  if we have already explored another node $M$ with a state sequence that prefixes $N$. Essentially, if the current goal being explored contains all the elements of a previously explored goal, then it shouldn't be further expanded. This is based on the observation that if we have already assessed the potential paths and outcomes for a specific goal, then exploring a more generalized version of the same goal is redundant. Such a mechanism avoids unnecessary repetitions, which streamlines the search process and improves the overall efficiency. Moreover, we define a valid proof as one that is devoid of ``cheating'' tactics (such as \textbf{sorry}) that tell Lean to assume that the current goal is completed, even though it has not been proven. This means that every path containing ``cheating'' tactics is disregarded. 

Errors in the search process typically manifest in two ways: a timeout or an exhaustion of nodes to search. We have allocated a three-minute window for each search, which is usually sufficient. We provided more analysis of the errors made by tactic generator in the experiment section.

\subsection{Result Intepreter}
If the correct answer is \emph{Unknown}, we only regard the result as correct if neither \emph{True} nor \emph{False} can be proven. All datasets investigated in this study only contain questions with only one correct answer. Consequently, if the proof system verifies more than one option, the response is immediately marked as incorrect.

\section{Experimental Setup}

We now describe our experimental setup: the datasets we used for evaluation and model training and the details of model training.

\subsection{Evaluation Data}
In our evaluation, we use two common logical reasoning datasets as testbeds:

\textbf{ProofWriter}: This deductive logical reasoning dataset presents problems in an intuitive language form. 
We incorporated the Open-World Assumption (OWA) subset as per \cite{logic-lm}, where each instance is characterized by a \{problem, goal\} pairing. The label for each pair contains \textsc{True}, \textsc{False}, or \textsc{Unknown}. It encompasses five segments based on the required reasoning depth.
Our focus is the depth-5 subset, which is the most challenging one. To get a fair comparison against Logic-LM, we used the same 600 sample tests, ensuring an even label distribution.

\textbf{FOLIO}: Unlike ProofWriter, FOLIO is constructed using first-order logic. This increases the complexity of the proving part. The dataset also presents problems in a more natural wording, with relationships that are considerably more complex. 
Such a combination of advanced logic and rich linguistic structure makes the formalization task in FOLIO substantially tougher than in ProofWriter. For our analysis, we turned to the entire FOLIO test set, encompassing 204 examples.

\subsection{Training Data for Domain Adaptation}
Regarding the data for model training, we collected 100 theorem proofs for ProofWriter and 27 theorem proofs for FOLIO, where each problem's proof was either manually annotated or collected from successful proofs generated by GPT-4. The data collection took about eight days.

During data annotation, we adopted two divergent approaches for constructing proofs. One approach emulated a straightforward strategy, encompassing a detailed procedure with all of the intermediate steps and lemmas, similar to how we humans might derive proof when given theorem-proving tasks. Conversely, the second approach resembles the proof formats found in mathlib.\footnote{\url{https://github.com/leanprover-community/mathlib}} We generate more succinct proofs of the same problem by reducing the number of intermediate lemmas and combining multiple tactics into a single compound tactic. The objective of having two annotations for the same problem was to examine the influence of annotation style on downstream logical reasoning. In the following experiments, we use \textbf{Intuitive} to refer to the first annotation style and \textbf{Concise} to denote the second annotation style. An illustrative example is available in Appendix \ref{app:c}.

It is important to mention that despite the limited data collected, the reasoning patterns for logical reasoning likely mirror those found in mathematical reasoning, which were potentially learned during pretraining. The main purpose of this data collection is domain adaptation to transfer from math to natural language logical reasoning.

\subsection{Model Training}
We used the same model structure for pretraining as in the ReProver paper, namely, Google's Byte-T5 \cite{byt5}. We also experimented with the pretrained ReProver from LeanDoJo \cite{leandojo}, which was pretrained on mathlib. The fine-tuning of our collected data took about six hours on one A100 40G. The hyperparameters are the same as in the original LeanDoJo paper. 


\section{Results}
\label{section:results}

We present our experimental results, including an examination of prompting-based baseline, experimental results for LeanReasoner, and a comparison between our work and other baselines.

\renewcommand{\arraystretch}{1.05}
\begin{table*}
    \centering
    \begin{tabular}{l|ccc|ccc}
        \hline
        \multicolumn{1}{c|}{\multirow{2}{*}{\textbf{Model}}} & \multicolumn{3}{c|}{\textbf{ProofWriter}} & \multicolumn{3}{c}{\textbf{FOLIO}} \\
        & \textbf{Formalize} & \textbf{Prove} & \textbf{Answer} & \textbf{Formalize} & \textbf{Prove} & \textbf{Answer} \\
        \hline
        GPT-4 Base & 94\% & 15\% & 80\% & 60\% & 10\% & 35\% \\
        GPT-4 Base Comments & 99\% & 15\% & 80\% & 75\% & 15\% & 35\% \\
        GPT-4 Base Separate & 95\% & 5\% & 75\% & 60\% & 10\% & 40\% \\
        \hline
        GPT-3 Base Comments & 77\% & 12\% & 63\% & 45\% & 10\% & 35\% \\
        \hline
        Logic-LM  & 98\% & 75.5\% & 74\% & 65\% & 69.2\% & 55\% \\
        \hline
    \end{tabular}
    \caption{Formalization, Proof, and Answer choice accuracy of 100 ProofWriter samples and 40 FOLIO samples via OpenAI Language Model API, with manual annotation.
    `GPT-4 Base' serves as our baseline, where few-shot examples include both formalization and proof generation in a single prompt. In `GPT-4 Base Comments', we augment these examples with line-by-line comments in Lean code. For `GPT-4 Base Separate', we divide the task into two parts, using separate prompts for formalization and proof generation.
    For simplicity, we did not use the self-refinement technique when evaluating Logic-LM.}
    \label{table1}
\end{table*}
\renewcommand{\arraystretch}{1.0}

\subsection{Prompting-Based Baselines}
\label{subsec:formalization}
Since there is no automated method to verify the accuracy of formalization, we conducted manual examinations of the formalized results to determine whether errors occur during formalization or proof generation stages.
In this examination, only formalizations that correctly capture every fact, axiom, and rule are counted as accurate. 
We manually examined 100 questions from ProofWriter's validation set and 40 questions from FOLIO's training set.
The findings have been summarized in \autoref{table1}.

\paragraph{Comparison of formalization accuracy.} The formalization accuracy of ProofWriter is much higher than FOLIO. This can be attributed to its simpler language structure. In the case of FOLIO, although using LLM for formalization helped in filtering out unnecessary details from the natural language context, there still exists some common error patterns. We have illustrated typical GPT-4 error patterns in Appendix \ref{app:b} using a composite sample derived from various error instances. Interestingly, Lean's formalization accuracy is on par with both Prolog and FOL in Logic-LM. This consistency underscores Lean's versatility, allowing it to uniformly represent different problem types within a single framework.

\paragraph{Adding textual comments increases formalization accuracy.} We observed improved results when formalized code was paired with descriptive textual comments (example in Appendix \ref{sec:proofwriterprompt}) sourced from the context. 
This approach further splits the formalization task into two subtasks: 1) linking textual context with formalized code and 2) generating formalized code based on the prior textual context. These textual cues acted as a bridge between raw text and formalized code, enhancing the performance of formalization.

\paragraph{GPT-3's performance on formalization is worse than GPT-4 .} The distinction in performance between GPT-3 and GPT-4 is evident. While the formalization for simpler problems is the same, GPT-3 struggles with intricate logic and complex problems. As such, we opted not to use GPT-3 in further tests. Additionally, we experimented with the CodeLLAMA \cite{codellama} model family for similar tasks, but found that their accuracy in formalization was significantly lower than that of GPT-3, achieving less than 30\% on ProofWriter.

\paragraph{The proof accuracy of prompting-based baseline is very low.}The proof accuracy section of the table is determined by whether the generated proof can be validated successfully in Lean. If the formalization of the question as a theorem is correct and the proof can be validated without any error or warning, then we can treat the proof as valid. However, the accuracy of rendered proofs is very low. The issue could stem from assigning too many tasks to the large language model, making it challenging to address both within a single prompt. Despite our efforts to separate formalization and proof, the results were still disappointing, which highlights GPT-3 and GPT-4's struggle with generating correct Lean proof. Interestingly, the proof accuracy of Logic-LM wasn't as high as we expected. Upon replicating their code, we found their chosen solver Pyke to be suboptimal, struggling to identify an answer when multiple search paths are available and some could result in loops.
\paragraph{The answer accuracy of prompting-based baseline is surprisingly high.} Despite the low accuracy in most of GPT-4's proofs, it achieved high accuracy for final choices on ProofWriter (as shown in column Answer). We believe this may be due to GPT-4's training exposure to the dataset, potentially leading to a degree of memorization.

\subsection{LeanReasoner}

\renewcommand{\arraystretch}{1.05}
\begin{table*}
  \setlength{\tabcolsep}{4pt}  

  \centering
  \begin{tabular}{l|c|c|ccc|ccc}
    \hline
    \multicolumn{1}{c|}{\multirow{3}{1.0cm}{\centering \textbf{Method}}} & 
    \textbf{Pretrained}
 & \textbf{Fine-tuned} &
    \multicolumn{3}{c}{\textbf{ProofWriter}} & \multicolumn{3}{|c}{\textbf{FOLIO}} \\
    & \textbf{on Math} & \textbf{on our } & \multicolumn{2}{c}{\textbf{Premise Selection}} & \textbf{Proof} & \multicolumn{2}{c}{\textbf{Premise Selection}} & \textbf{Proof} \\

     & \textbf{ Data} & \textbf{Annotation} & Rec@1 & Rec@4 & Acc & Rec@1 & Rec@4 & Acc \\
     \hhline{---------}
    GPT-4 & N/A & N/A & \multicolumn{2}{c}{N/A} & 15\% & \multicolumn{2}{c}{N/A} & 10\% \\
    LeanReasoner & Yes & No & 56.2\% & 81.3\% & 0\% & 23.5\%  & 38.2\%        & 0\% \\
    LeanReasoner & No & Intuitive & 62.5\%  & 100\% & 99\% & 54.8\% & 95.2\% & 71.4\% \\
    LeanReasoner & Yes & Intuitive & 75\% & 100\% & 99\% & 71.4\% & 96.8\% & 85.7\%\\
    LeanReasoner & Yes & Concise & 75\% & 100\% & 99\% & 83.8\% & 97.4\% & 85.7\%\\
    \hline
  \end{tabular}
  \caption{Comparative Analysis of Recall@k in premise selection and overall proof accuracy for 99 ProofWriter test samples and 28 FOLIO test samples. Note the proof accuracy here is different from Table \ref{table1} because it is directly linked to the final accuracy. The effects of pretraining and fine-tuning on LeanReasoner are evaluated using theorem-proving data and both Intuitive and Concise annotation sets, respectively. Premise Selection accuracy was not calculated for the GPT-4 baseline due to the complexities in prompting GPT-4 with Lean goals.}\label{tab:agreement}
\end{table*}
\renewcommand{\arraystretch}{1.0}

In this section, we focus on training our own models to do tactic generation using our annotated training data. To isolate the impact of erroneous formalization, we only used the accurate formalizations from the previous subsection for testing. This gave us 99 test examples for ProofWriter and 28 for FOLIO. All findings are detailed in Table \ref{tab:agreement}.

\paragraph{Fine-tuning on annotated data increases premise selection accuracy.} We first compare the results on premise selection using the metrics recall@1 and recall@4. The recall@k metric is defined as follows: \[ \text{recall@k} = \frac{\left| \text{GT\_Prem} \cap \text{Pred\_Prem}{[0:k]} \right|}{\left| \text{GT\_Prem} \right|} \text{,}\] where {GT\_Prem} means ground truth premises and {Pred\_Prem} means top predicted premises. The suboptimal results of LeanReasoner pretrained solely with math data may be attributed to the domain mismatch between mathematical theorem proving and logical reasoning. The model frequently makes mistakes by attempting to use other, unrelated tactics that are useful in mathematical theorem proving (like \textbf{ring}, \textbf{linarith}) but not applicable in logical reasoning.
Furthermore, the accuracy for FOLIO was noticeably poorer than that of ProofWriter. This disparity is likely due to FOLIO's intricate logic and its need for a broader array of first-order logic tactics such as \textbf{cases}, \textbf{have}, and \textbf{contradiction}. In contrast, ProofWriter primarily employs tactics like \textbf{apply}, \textbf{exact}, and \textbf{split}.

\paragraph{Pretraining on theorem proving data increases overall accuracy.} Regarding the overall proof results, LeanReasoner pretrained on math theorem proving data consistently outperformed other approaches for both ProofWriter and FOLIO datasets. This success indicates that our model effectively uses the logical ``nuggets'' found in mathematical theorem proofs. While the premise selector benefits from distinct cues and a limited range of choices, the realm of tactic generation is much broader. This vastness of options renders the ReProver baseline's proof accuracy nearly negligible. But other than that, there is a strong correlation between premise selection accuracy and overall proof accuracy. While the benefits of a pretrained LeanReasoner may not be as noticeable for simpler datasets like ProofWriter, its value becomes evident for more complex datasets, such as FOLIO.

\begin{figure*}
\centering
\includegraphics[trim=0.9cm 0cm 2.1cm -2.9cm, width=1\textwidth]{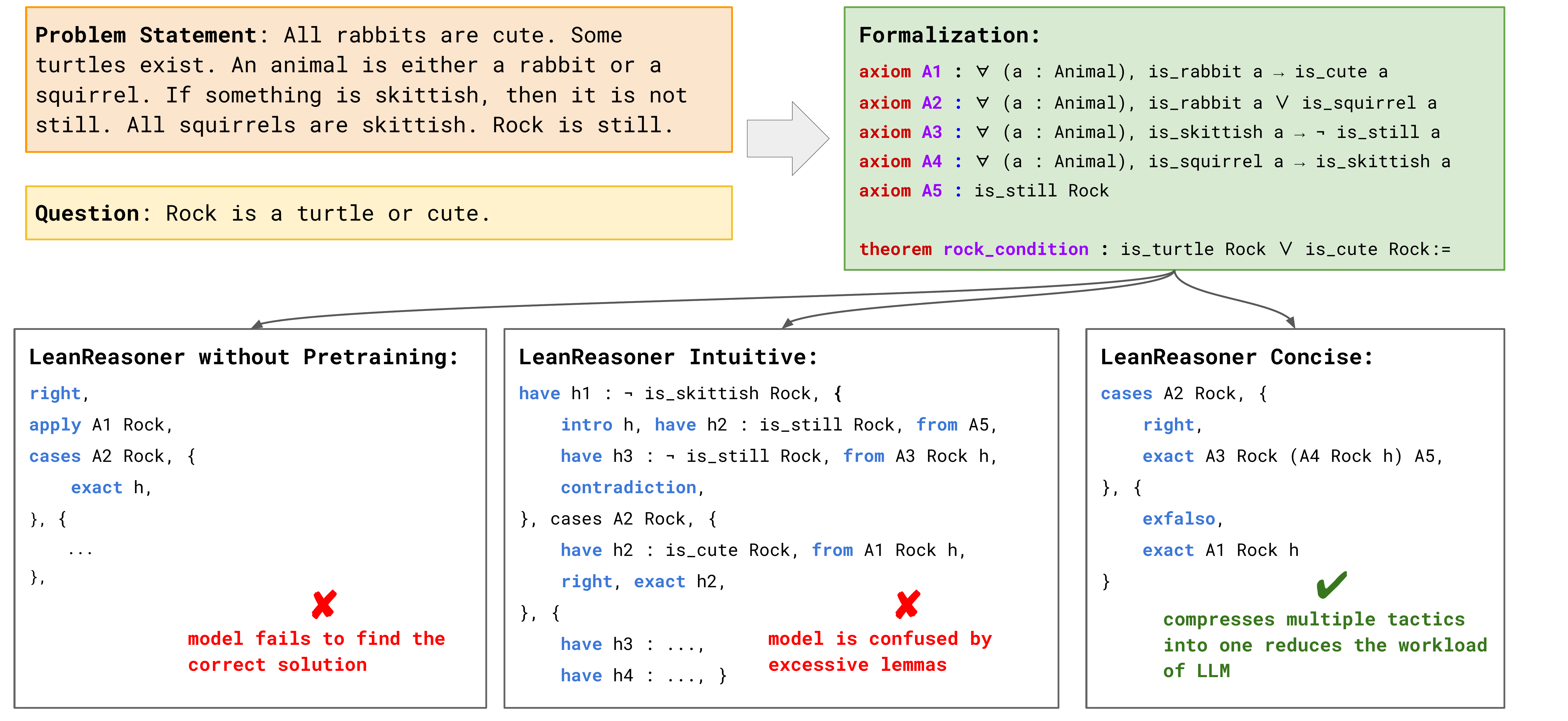} 
\caption{Sample proofs created by LeanReasoner without pretraining (\textbf{left}), finetuned on \textbf{Intuitive} data (\textbf{middle}), and finetuned on \textbf{Concise} data (\textbf{right}).}
\label{case_study}
\end{figure*}

\paragraph{\textbf{Concise} annotation gives better result on premise selection.}Fine-tuning with different annotations has a slight effect on premise selection and tactic generation in this small test set. When fine-tuned with \textbf{Concise} annotations, LeanReasoner would also try to generate concise proofs, which usually use compound tactics that offer more information for premise selection. However, the final proof accuracy has not changed on this small test set. \autoref{case_study} displays an example of proofs for the same question, produced by the three primary methods we compared. In the absence of pretraining, the model struggles to identify an appropriate approach for solving the problem. It merely attempts to apply the next applicable theorem, lacking a clear objective. While \textbf{Intuitive} data offers numerous lemmas that assist in the thought process during proof-writing, these excessive lemmas do not aid LLMs in generating tactics effectively.


\subsection{Other Baselines}
\renewcommand{\arraystretch}{1.05}
\begin{table}[]
    \centering
    \begin{tabular}{lc}
    \hline
        \textbf{Method} & \textbf{Acc}   \\ \hline
        \multicolumn{2}{l}{\textbf{Full training set method}} \\
        \hline
        Abs Biases \cite{induction} & 80.6\% \\
        MetaInduce \cite{MetaInduce} & 98.6\%\\
        RECKONING \cite{RECKONING} & 99.8\%\\ \hline
        \multicolumn{2}{l}{\textbf{Zero-shot method}} \\
        \hline
        GPT-4 CoT \cite{logic-lm} & 68.1\% \\ 
        Logic-LM \cite{logic-lm} & 79.3\% \\ \hline
        \multicolumn{2}{l}{\textbf{Our method (finetuned on 100 samples)}} \\
        \hline
        LeanReasoner without Pretraining  & 95.8\% \\ 
        LeanReasoner fine-tuned on Intuitive   & 98.3\% \\ 
        LeanReasoner fine-tuned on Concise & 98.3\%  \\ \hline
    \end{tabular}
    \caption{The fine-tuned LeanReasoner has been pretrained on mathlib. Full training set method means the model has been trained on the full training set of ProofWriter. \textbf{Fine-tuning on Concise achieves near-perfect accuracy with significantly less data}.}
    \label{tab:comparsion_proofwriter}
\end{table}
\renewcommand{\arraystretch}{1} 

\renewcommand{\arraystretch}{1.05}
\begin{table}[]
    \centering
    \begin{tabular}{lc}
    \hline
        \textbf{Method} & \textbf{Acc}  \\ \hline
                \multicolumn{2}{l}{\textbf{Full training set method}} \\
        \hline
        Roberta \cite{folio} & 62.1\% \\
        FOLNet \cite{FOLNet} & 70.6\% \\
        \hline
        \multicolumn{2}{l}{\textbf{Zero-shot method}} \\
        \hline
        GPT-4 CoT \cite{logic-lm} & 70.6\% \\ 
        Logic-LM \cite{logic-lm} & 74.5\% \\
        
        Lean Z3 (SATLM)  & 77.5\% \\ 
        \hline
        \multicolumn{2}{l}{\textbf{Our method (finetuned on 27 samples)}} \\
        \hline
        LeanReasoner without Pretraining  & 66.2\% \\ 
        LeanReasoner fine-tuned on Intuitive  & 78.4\% \\ 
        LeanReasoner fine-tuned on Concise  & 82.6\% \\ \hline
    \end{tabular}
    \caption{Result from ``Lean Z3'' is derived from lean-smt applied to formalized Lean Code. The fine-tuned LeanReasoner has been pretrained on mathlib. Full training set method means the model has been trained on the full training set of FOLIO. \textbf{Our approach achieves state-of-the-arts performance on FOLIO}.}
    \label{tab:comparsion_FOLIO}
\end{table}
\renewcommand{\arraystretch}{1.0}

Having demonstrated that pretraining on theorem-proving data yields superior performance, we proceed to benchmark our results against established baselines for both ProofWriter and FOLIO. The evaluation uses the same set of 600 problems from LogicLM and the entire FOLIO test set. 

\paragraph{Our approach yields near-perfect accuracy on ProofWriter with significantly less data.}As illustrated in Table~\ref{tab:comparsion_proofwriter}, our approach yields near-perfect accuracy on the ProofWriter dataset. While other methods except Logic-LM and GPT-4 COT use the entire training set of ProofWriter, our approach relies on just 100 examples, underscoring the efficiency of our method. Fine-tuning on \textbf{Concise} annotation does not bring any advantage to the final performance on this dataset.

\paragraph{Our approach achieves state-of-the-arts performance on FOLIO.} \autoref{tab:comparsion_FOLIO} presents our performance on FOLIO. For a fair comparison with SATLM that uses the Z3 solver, we used the lean-smt tool\footnote{https://github.com/ufmg-smite/lean-smt} on our formalized Lean code. This tool produces outcomes in the form of ``sat/unsat''. In Z3, ``sat'' stands for ``satisfiable.'' When Z3 returns ``sat'' as the result, it means that there exists a set of variable values that makes the theorem true. On the other hand, ``unsat'' stands for ``unsatisfiable''. When Z3 returns ``unsat'', it means that the formula is inherently contradictory and cannot be satisfied under any circumstance. We interpret these results similarly to ``found a proof/didn't find a proof'' using our result interpreter. Due to the extensive length of proofs for FOLIO problems, we observed that when LeanReasoner is fine-tuned on the \textbf{Intuitive} dataset, it often allocates an excessive amount of time for exploration and occasionally enters loops. In contrast, generating shorter proofs tends to ease the discovery of the proof. While the tactics generated when fine-tuned on the \textbf{Concise} dataset are more challenging to produce, the bottleneck for LeanReasoner on FOLIO resides in the search process.

\paragraph{Challenges in Benchmarking.} It is important to acknowledge that there can be scenarios where errors in problem formalization or proof generation may occur, yet the final answer is still deemed correct. A case in point is when the answer to a problem is $Unknown$, and errors arise in these stages. In such instances, the model would struggle to prove either the positive or negative theorem. However, with our result interpreter, these instances would still be classified as correct despite the underlying issues in problem handling.


\section{Related Work}
\paragraph{Combining LLM with symbolic solver.}Several past studies \cite{FOLNet,faithful,RECKONING} used symbolic solvers to augment neural networks with logical reasoning. Many of these approaches have limitations, like the necessity for custom or specialized module designs that lack broad applicability. Recent work \cite{logic-lm,satlm,certified,linc} presents a more general framework that combines contemporary LLMs with symbolic logic, bypassing the need to train or craft intricate modules tailored for specific problems. While our research aligns with these, we do not exclusively rely on off-the-shelf solvers.

\paragraph{Boosting the reasoning skill of LLM by training on reasoning data.}A common way to boost the reasoning skills of LLMs is by training them on data that requires some form of reasoning. As noted by \newcite{quantitive_llm}, LLMs trained with science and math data do better on tasks that require reasoning, especially when using CoT prompting. Other results by \newcite{fu2022gptroadmap} and \newcite{complexity} suggest that powerful LLMs obtain advanced reasoning capabilities from being trained on code. This work is an extension of this idea to theorem proving.


\section{Conclusion}
We introduced LeanReasoner, a framework based on Lean that augments the logical reasoning abilities of LLMs. We follow an extensive examination of errors from the formalization and proof generation stages that are present in our framework. We also examine the performance enhancements from pretraining on theorem-proving data and annotation styles (concise v.s. intuitive). We offered a comprehensive comparison with other techniques that highlight the strengths of our model.
Our results underscore the potential of integrating theorem-proving frameworks with LLMs in advancing logical reasoning.


\section*{Limitations}
Despite our promising results, our method encounters limitations when dealing with problems that involve commonsense and factual reasoning. In these cases, it is challenging to retrieve all the necessary information and accurately represent it in Lean. Consider MMLU \cite{mmlu} and SummEdits \cite{summedits}: MMLU requires the model to possess extensive world knowledge, while SummEdits involves determining consistency in summaries of different edits. In both instances, the ability to represent the complexity and nuance of real-world knowledge in Lean is severely limited. 

Further complications arise when dealing with math word problems \cite{gsm8k} and similar tasks \cite{math}, where the goal is to derive a numeric solution rather than a proof. The theorem-proving approach, while effective for certifying the validity of logical reasoning, does not directly yield a numerical answer. Lastly, our method grapples with problems found in more complicated reasoning datasets like TheoremQA \cite{theoremqa}. These problems require an advanced understanding of complex concepts and the ability to formalize these concepts into Lean. Our current framework struggles with this level of complexity, underscoring the need for more sophisticated formalization techniques and a deeper integration between language understanding and theorem proving.

Even in the context of symbolic problems, there are challenges. For instance, consider the LogicalDeduction task from the BigBench dataset \cite{bigbench}. Although this problem appears straightforward, employing Lean to solve it is neither the most practical nor the most efficient approach. Lean, as a theorem prover, is excellent in abstract reasoning and proof construction, but when faced with tasks involving constraints and variable possibilities, it falls short. To solve the problems in LogicDeduction, using Lean would require us to formalize the concepts of ordering and relative positioning. Even after doing so, generating proof would necessitate significant labor and wouldn't necessarily yield a readily interpretable answer. In contrast, a Constraint Satisfaction Problem (CSP) solver can effectively manage constraints and generate potential solutions efficiently.

\section*{Ethical Considerations}

Incorporating Lean's theorem-proving capabilities into LLMs offers a layer of mathematical rigor that improves the reliability of conclusions derived. However, LLMs are known to be susceptible to data biases, which may manifest in critical applications. This issue can inadvertently lead to skewed logic or unintended bias in sensitive domains such as medical diagnoses or legal interpretations. While our method's foundation in Lean's theorem proving data acts as a rigorous check, complete reliance on it is not foolproof. A proactive approach in reviewing both training data and model outcomes is essential to uphold unbiased reasoning.


\section*{Acknowledgements}

We thank the reviewers and members of the Cohort for the valuable feedback and comments on the paper. We appreciate the provision of computing resources through the Baskerville cluster at the University of Birmingham.

\newpage

\bibliography{anthology,custom}
\bibliographystyle{acl_natbib}

\appendix

\newpage
\section{Prompts for Formalization}
\subsection{Prompts for ProofWriter}
\label{sec:proofwriterprompt}
In subsection \ref{subsec:formalization}, we discussed various formalization approaches. In this section, we present the results using the GPT-4 Base Comments method on ProofWriter when the answer is False. As evident from the last line, the predicted outcome from GPT-4 can be derived easily.

\begin{lstlisting}
(*@\color{codepurple}{\textbf{System Message}}@*): 
You are a logician with a background in 
mathematics that translates natural language 
reasoning text to Lean code so that these natural 
language reasoning problems can be solved. During 
the translation, please pay close attention to the 
predicates and entities. There is an additional 
requirement: I also want you to try to prove the 
theorem you translated to Lean. If you can prove the 
theorem, give me True at the end of the answer. If 
you can prove the negation of the theorem, write 
False at the end of the answer. If you can neither
prove the original theorem nor the negation 
of the theorem, please give me Unknown at the end of
the answer.
\end{lstlisting}

\begin{lstlisting}
(*@\color{codepurple}{\textbf{Input}}@*):
Textual context: The cat is blue. The cat is nice.
The cat likes the cow. The cat needs the cow. The
cat needs the rabbit. The cow does not need the 
rabbit. The cow needs the tiger. The rabbit is 
round. The tiger is nice. The tiger does not 
like the cow. If someone likes the cat and they
chase the cat then they are blue. If someone
likes the cow and they are red then the cow is 
round. If someone needs the tiger and they need 
the cat then they do not chase the cow. If someone
needs the cat and the cat is blue then the cat is
red. If someone is round then they need the cat. 
If someone likes the cat and the cat needs the 
rabbit then the cat chases the tiger. If the rabbit
needs the tiger then the rabbit is not round.

Question: Based on the above information, is the 
following statement true, false, or unknown? The 
cow chases the cow.
\end{lstlisting}

\newpage

\begin{lstlisting}[language=lean]
(*@\color{codepurple}{\textbf{Output}}@*):
universe u

constant obj : Type u

constant Cat : obj
constant Cow : obj
constant Rabbit : obj
constant Tiger : obj

constant Blue : obj → Prop
constant Nice : obj → Prop
constant Likes : obj → obj → Prop
constant Needs : obj → obj → Prop
constant Round : obj → Prop
constant Red : obj → Prop
constant Chases : obj → obj → Prop

axiom T1 : Blue Cat
axiom T2 : Nice Cat
axiom T3 : Likes Cat Cow
axiom T4 : Needs Cat Cow
axiom T5 : Needs Cat Rabbit
axiom T6 : ¬ Needs Cow Rabbit
axiom T7 : Needs Cow Tiger
axiom T8 : Round Rabbit
axiom T9 : Nice Tiger
axiom T10 : ¬ Likes Tiger Cow

-- If someone likes the cat and they chase the cat then they are blue
axiom R1 : ∀ x : obj, Likes x Cat ∧ Chases x Cat → Blue x
-- If someone likes the cow and they are red then the cow is round
axiom R2 : ∀ x : obj, Likes x Cow ∧ Red x → Round Cow
-- If someone needs the tiger and they need the cat then they do not chase the cow
axiom R3 : ∀ x : obj, Needs x Tiger ∧ Needs x Cat → ¬ Chases x Cow
-- If someone needs the cat and the cat is blue then the cat is red
axiom R4 : ∀ x : obj, Needs x Cat ∧ Blue Cat → Red Cat
-- If someone is round then they need the cat
axiom R5 : ∀ x : obj, Round x → Needs x Cat
-- If someone likes the cat and the cat needs the rabbit then the cat chases the tiger
axiom R6 : ∀ x : obj, Likes x Cat ∧ Needs Cat Rabbit → Chases Cat Tiger
-- If the rabbit needs the tiger then the rabbit is not round
axiom R7 : ∀ x : obj, Needs Rabbit Tiger → ¬ Round Rabbit


theorem cow_chases_cow : Chases Cow Cow :=
begin
  have H1 : Needs Cow Tiger := T7,
  have H2 : Needs Cat Cow := T4,
  have H3 : ¬ Chases Cow Cow := R3 Cow (and.intro H1 H2),
  contradiction,
  sorry,
end

-- Looks like the theorem cannot be proven in Lean.
-- So let's try to prove the negation of the theorem, which is Cow does not chase Cow

theorem not_cow_chases_cow : ¬ Chases Cow Cow :=
begin
  apply R3 Cow,
  split,
  exact T7,
  apply R5,
  apply R2 Cat,
  split,
  exact T3,
  apply R4 Rabbit,
  split,
  apply R5,
  exact T8,
  exact T1,
end

-- The negation of the theorem can be proven in Lean. So
-- The answer is False

\end{lstlisting}

\subsection{Prompts Used for FOLIO}
\label{sec:FOLIOprompt}

For FOLIO, the prompts differ slightly from those used for ProofWriter. Since FOLIO's textual context remains consistent across multiple questions, we concatenate questions sharing the same context and prompt the large language model in a single instance. An illustrative example is provided below.

\begin{lstlisting}
(*@\color{codepurple}{\textbf{System Message}}@*): 
You are a logician with a background in 
mathematics that translates natural language 
reasoning text to Lean code so that these natural 
language reasoning problems can be solved. During 
the translation, please pay close attention to the 
predicates and entities. There is an additional 
requirement: I also want you to try to prove the 
theorem you translated to Lean. If you can prove the 
theorem, give me True at the end of the answer. If 
you can prove the negation of the theorem, write 
False at the end of the answer. If you can neither
prove the original theorem nor the negation 
of the theorem, please give me Unknown at the end of
the answer.
\end{lstlisting}

\begin{lstlisting}
(*@\color{codepurple}{\textbf{Input}}@*):
Textual context: There are six types of wild 
turkeys: Eastern wild turkey, Osceola wild 
turkey, Gould’s wild turkey, Merriam’s wild 
turkey, Rio Grande wild turkey, and Ocellated
wild turkey. Tom is not an Eastern wild turkey.
Tom is not an Osceola wild turkey. Tom is
also not a Gould's wild turkey, or a 
Merriam's wild turkey, or a Rio Grande wild
turkey. Tom is a wild turkey.

Question 1: Based on the above information, is the 
following statement true, false, or uncertain? 
Tom is an Ocellated wild turkey.
Question 2: Based on the above information, is the
following statement true, false, or uncertain? 
Tom is an Eastern wild turkey.
Question 3: Based on the above information, is the
following statement true, false, or uncertain? 
Joey is a wild turkey.
\end{lstlisting}

\newpage

\begin{lstlisting}[language=lean]
(*@\color{codepurple}{\textbf{Output}}@*):
-- Introducing basic types for entities in our universe.
constant Turkey : Type
constant Tom : Turkey
constant Joey : Turkey

-- Defining properties for the turkeys in our universe.
constant is_wild_turkey : Turkey → Prop
constant is_eastern_wild_turkey : Turkey → Prop
constant is_osceola_wild_turkey : Turkey → Prop
constant is_goulds_wild_turkey : Turkey → Prop
constant is_merriams_wild_turkey : Turkey → Prop
constant is_rio_grande_wild_turkey : Turkey → Prop
constant is_ocellated_wild_turkey : Turkey → Prop

-- Context from the information

-- Tom is not an Eastern wild turkey.
axiom A1 : ¬ is_eastern_wild_turkey Tom
-- Tom is not an Osceola wild turkey.
axiom A2 : ¬ is_osceola_wild_turkey Tom
-- Tom is also not a Gould's wild turkey.
axiom A3 : ¬ is_goulds_wild_turkey Tom
-- Tom is not a Merriam's wild turkey.
axiom A4 : ¬ is_merriams_wild_turkey Tom
-- Tom is not a Rio Grande wild turkey.
axiom A5 : ¬ is_rio_grande_wild_turkey Tom
-- Tom is a wild turkey.
axiom A6 : is_wild_turkey Tom
-- There are six types of wild turkeys: Eastern wild turkey, Osceola wild turkey, Gould’s wild turkey, Merriam’s wild turkey, Rio Grande wild turkey, and Ocellated wild turkey.
axiom turkey_type_exclusivity : ∀ (t : Turkey),
    is_wild_turkey t → is_eastern_wild_turkey t ∨ is_osceola_wild_turkey t ∨ is_goulds_wild_turkey t ∨ is_merriams_wild_turkey t ∨ is_rio_grande_wild_turkey t ∨ is_ocellated_wild_turkey t

-- Formalizing the Questions

-- Question 1
-- Tom is an Ocellated wild turkey.
-- Let's first prove the positive case.
theorem tom_is_ocellated_wild_turkey : is_ocellated_wild_turkey Tom :=
begin
    cases turkey_type_exclusivity Tom A6, {
        have h1 : ¬ is_eastern_wild_turkey Tom, from A1,
        contradiction,
    }, {
        cases h, {
            have h1 : ¬ is_osceola_wild_turkey Tom, from A2,
            contradiction,
        }, {
            cases h, {
                have h1 : ¬ is_goulds_wild_turkey Tom, from A3,
                contradiction,
            }, {
                cases h, {
                    have h1 : ¬ is_merriams_wild_turkey Tom, from A4,
                    contradiction,
                }, {
                    cases h, {
                        have h1 : ¬ is_rio_grande_wild_turkey Tom, from A5,
                        contradiction,
                    }, {
                        assumption,
                    }
                }
            }
        }
    }
end

-- Let's try the negative case.

theorem not_tom_is_ocellated_wild_turkey : ¬ is_ocellated_wild_turkey Tom :=
begin
    exact A2,
end

-- The answer is Unknown

-- Question 2
-- Tom is an Eastern wild turkey.
-- Let's first prove the positive case.
theorem tom_is_eastern_wild_turkey : is_eastern_wild_turkey Tom :=
begin
    sorry, -- Placeholder for proof
end

-- Let's try the negative case.

theorem not_tom_is_eastern_wild_turkey : ¬ is_eastern_wild_turkey Tom :=
begin
    exact A1,
end

-- The answer is False

-- Question 3
-- Joey is a wild turkey.
-- Let's first prove the positive case.
theorem joey_is_wild_turkey : is_wild_turkey Joey :=
begin
    sorry, -- Placeholder for proof
end

-- Let's try the negative case.

theorem not_joey_is_wild_turkey : ¬ is_wild_turkey Joey :=
begin
    sorry, -- Placeholder for proof
end

-- The answer is Unknown
\end{lstlisting}

\section{Example Errors During Formalization}
\label{app:b}
This is an artificial example highlighting typical errors encountered in the formalization stage.
\begin{lstlisting}
(*@\color{codepurple}{\textbf{Input}}@*): 
There are three types of wild turkeys: Eastern
turkey, Osceola turkey, Gould’s turkey. Tom is 
neither an Eastern turkey nor an Osceola turkey.
Tom is a wild turkey. Bob is a Eastern turkey. 
Bob weights 86 pounds and Tom weights 73 pounds.
Turkey either bites or stings.Bob bites. Eastern 
turkey 's bite would cause death.   

Question 1:  Tom is a Gould's turkey. 
Question 2:  Bob is heavier than Tom. 
Question 3:  If Bob bites you, you will die.
\end{lstlisting}

\begin{lstlisting}[language=lean]
(*@\color{codepurple}{\textbf{Output}}@*):
constant Turkey : Type
constant Eastern Osceola Goulds : Turkey → Prop
constant weight_86 weight_73 : Turkey → Prop
constant bites stings : Turkey → Prop
constant eastern_bite_causes_death : Turkey → Prop
constant Tom Bob : Turkey

axiom A1 : ¬ Eastern Tom ∧ ¬ Osceola Tom
axiom A2 : Eastern Bob
axiom A3 : weight_86 Bob
axiom A4 : weight_73 Tom
axiom A5 : ∀ (t : Turkey), bites t ∨ stings t
axiom A6 : bites Bob
axiom A7 : ∀ (t : Turkey), eastern_bite_causes_death t
\end{lstlisting}

\newpage
There are a few errors in the above code, including:

\begin{itemize}
\item  There is a missing axiom that corresponds to "There are three types of wild turkeys": 
\item[] {\small Goulds Tom $\vee$ Eastern Tom $\vee$  Osceola Tom}
\item  The formalization of numbers is incorrect, it should be: 
\item[] {\small constant weight : Turkey $\to$ $\mathbb{N}$}
\item[] {\small axiom A3 : weight Bob = 86}
\item[] {\small axiom A4 : weight Tom = 73}
\item  The formalization of logic is incorrect, it should be: 
\item[] {\small $\neg$ bite\_causes\_death t $\wedge$ bite\_causes\_itching t) $\vee$ (bite\_causes\_death $\wedge$ $\neg$ bite\_causes\_itching t}
\item  There is an incorrect division of concepts that would make the proving impossible, the correct version should be:
\item[] {\small $\forall$ (t : Turkey), Eastern t $\rightarrow$ bite\_causes\_death t}
\end{itemize}

\section{Example Proof Annotation with Different Annotation Styles}
\label{app:c}
Here we're showing two example proofs created on the same problem with `Intuitive' annotation style and `Concise' annotation style.

\begin{lstlisting}
(*@\color{codepurple}{\textbf{Input}}@*): 
"Textual Context": All eels are fish. No fish are
plants. A thing is either a plant or animal. 
Nothing that breathes is paper. All animals breathe.
If a sea eel is either an eel or a plant, then a sea
eel is an eel or an animal.
"Question": "Based on the above information, is 
the following statement true, false, or uncertain?
Sea eel is a paper.
\end{lstlisting}

\begin{lstlisting}[language=lean]
(*@\color{codepurple}{\textbf{Formalized Context}}@*):
constant Thing : Type

constant is_eel : Thing → Prop
constant is_fish : Thing → Prop
constant is_plant : Thing → Prop
constant is_animal : Thing → Prop
constant is_paper : Thing → Prop
constant breathes : Thing → Prop

constant sea_eel : Thing

-- All eels are fish.
axiom A1 : ∀ (t : Thing), is_eel t → is_fish t
-- No fish are plants.
axiom A2 : ∀ (t : Thing), is_fish t → ¬ is_plant t
-- A thing is either a plant or animal.
axiom A3 : ∀ (t : Thing), is_plant t ∨ is_animal t
-- Nothing that breathes is paper.
axiom A4 : ∀ (t : Thing), breathes t → ¬ is_paper t
-- All animals breathe.
axiom A5 : ∀ (t : Thing), is_animal t → breathes t
-- If a sea eel is either an eel or a plant, then a sea eel is an eel or an animal.
axiom A6 : (is_eel sea_eel ∨ is_plant sea_eel) → (is_eel sea_eel ∨ is_animal sea_eel)
\end{lstlisting}

\begin{lstlisting}[language=lean]
(*@\color{codepurple}{\textbf{Intuitive Proof}}@*):
theorem not_sea_eel_is_paper : ¬ is_paper sea_eel :=
begin
    cases A3 sea_eel, {
        have h1 : ¬ is_fish sea_eel, {
            intro h,
            have temp := A2 sea_eel h,
            contradiction,
        },
        have h2 : ¬ is_eel sea_eel, {
            intro h,
            have temp := A1 sea_eel h,
            contradiction,
        },
        have h3 : is_eel sea_eel ∨ is_plant sea_eel, {
            right,
            assumption,
        },
        have h4 : is_eel sea_eel ∨ is_animal sea_eel := A6 h3,
        cases h4, {
            contradiction,
        }, {
            have h5 : breathes sea_eel := A5 sea_eel h4,
            have h6 : ¬ is_paper sea_eel := A4 sea_eel h5,
            contradiction,
        }
    }, {
        have h1 : breathes sea_eel := A5 sea_eel h,
        have h2 : ¬ is_paper sea_eel := A4 sea_eel h1,
        contradiction,
    }
end
\end{lstlisting}

\begin{lstlisting}[language=lean]
(*@\color{codepurple}{\textbf{Concise Proof}}@*):
theorem not_sea_eel_is_paper : ¬ is_paper sea_eel :=
begin
    cases A3 sea_eel, {
        cases A6 (or.inr h), {
            have h1 := A2 sea_eel (A1 sea_eel h_1),
            contradiction,
        }, {
            exact A4 sea_eel (A5 sea_eel h_1),
        }
    }, {
        exact A4 sea_eel (A5 sea_eel h),
    }
end
\end{lstlisting}





\end{document}